\documentclass{article}
\PassOptionsToPackage{numbers,compress}{natbib}
\usepackage[preprint]{neurips_2026}
\usepackage[utf8]{inputenc}
\usepackage[T1]{fontenc}
\usepackage{hyperref}
\hypersetup{hidelinks}
\usepackage{url}
\usepackage{booktabs}
\usepackage{amsmath,amsfonts}
\usepackage{microtype}
\usepackage[table]{xcolor}
\usepackage{colortbl}
\usepackage{graphicx}
\usepackage{subcaption}
\usepackage{multirow}
\newcommand{\method}{CortexBridge}
\definecolor{rowblue}{HTML}{E6F0FA}
\definecolor{gainfg}{HTML}{2E7D32}
\newcommand{\gain}[1]{\textcolor{gainfg}{\scriptsize{$\uparrow$#1}}}
\newcommand{\loss}[1]{\textcolor{red!70}{\scriptsize{$\downarrow$#1}}}

\newif\ifanonymoussubmission
\anonymoussubmissionfalse

\title{CortexBridge: Cortical Alignment of EEG Montages for Foundation Models}
\ifanonymoussubmission
  \author{Anonymous Authors}
\else
  \author{
  Jiazhen Hong$^{1}$ \quad
  Xiaotian Zhou$^{1}$ \quad
  Zihao Ding$^{1}$ \quad
  Kailong Wang$^{1}$ \quad
  Yu Wu$^{1}$ \\[0.5em]
  $^{1}$Department of Electrical and Computer Engineering, Rutgers University \\
  Piscataway, NJ 08854, USA
  }
\fi

\begin{document}
\maketitle

\begin{abstract}
Electroencephalography (EEG) foundation models are often pretrained with a fixed channel vocabulary or a limited set of montages, making transfer difficult when electrode layouts change. We propose \method, a lightweight adapter that combines EEG features with electrode and atlas coordinates to map arbitrary montages into a shared cortical latent space. Evaluated with three frozen foundation models on five brain-computer interface (BCI) datasets from the Mother of All BCI Benchmarks (MOABB), \method{} improves performance in 13 of 15 evaluations. The gains in balanced accuracy average 0.80\% for EEGPT, 0.70\% for LaBraM, and 3.26\% for CBraMod, with a maximum gain of 13.02\% on 12-class steady-state visual evoked potential (SSVEP) classification. Visualizations of the learned atlas representations reveal task-dependent spatial patterns, with SSVEP showing a more concentrated representation in the Yeo Visual network than auditory P300. These results establish cortical alignment as a learnable and anatomically grounded routing mechanism from heterogeneous EEG montages to pretrained foundation models.
\end{abstract}

\section{Introduction}

Recent advances in EEG representation learning have shifted the field from task-specific decoders toward self-supervised pretraining and large-scale foundation models~\citep{kostas2021bendr,chien2022maeeg,mohammadi2024eeg2rep,jiang2024labram,wang2024eegpt,wang2024cbramod}. Despite this shift, most models still represent EEG in scalp-electrode space. Their spatial interfaces are therefore tied to the acquisition montage rather than to a shared anatomical representation.

A key challenge is montage dependence: electrode number and placement vary across recordings, while many pretrained models rely on fixed or predefined channel interfaces~\citep{jiang2024labram,wang2024eegpt,wang2024cbramod}. Coordinate-based interfaces relax channel identity constraints but still operate in scalp space~\citep{hong2025samba, fang2025neuript,ouahidi2025reve}. Classical source imaging maps scalp measurements to anatomy, but relies on explicit forward and inverse modeling~\citep{michel2019eegsource,pascualmarqui2002sloreta}. This motivates cortical alignment as a shared representation across heterogeneous montages for foundation models, distinct from biophysical source estimation.

We propose \method, a lightweight adapter that combines EEG features with electrode and atlas coordinates to map arbitrary montages into a shared cortical latent space. \method{} learns sparse electrode-to-ROI routing, models multi-scale interactions among cortical regions, and injects the resulting representation into a frozen foundation model through a residual interface. Evaluated with three frozen foundation models on five MOABB datasets, \method{} improves performance in 13 of 15 evaluations.

Our contributions are:
\begin{itemize}

\item We introduce cortical alignment as a learnable interface for adapting heterogeneous EEG montages to pretrained foundation models.

\item We develop a lightweight adapter that combines signal content with electrode and atlas geometry for sparse region-of-interest (ROI) routing.

\item We demonstrate consistent gains across three frozen foundation models, five public datasets, and three EEG paradigms, while learned atlas representations reveal task-dependent spatial organization.
\end{itemize}

\section{Related Work}

\paragraph{EEG representation learning.}
Early EEG models such as EEGNet and EEG Conformer learn task-specific spatial and temporal representations~\citep{lawhern2018eegnet,song2022eeg}. Self-supervised methods including BENDR, MAEEG, and EEG2Rep instead learn reusable representations through contrastive or masked objectives~\citep{kostas2021bendr,chien2022maeeg,mohammadi2024eeg2rep}. More recent foundation models broaden pretraining scale and architecture: MMM incorporates geometric structure~\citep{yi2023learning}, BIOT uses frequency-aware biosignal tokens~\citep{yang2023biot}, LaBraM predicts masked neural codes~\citep{jiang2024labram}, EEGPT aligns spatial and temporal representations~\citep{wang2024eegpt}, and CBraMod models spatial and temporal dependencies in parallel~\citep{wang2024cbramod}.

\begin{figure}
\centering
\includegraphics[width=\linewidth]{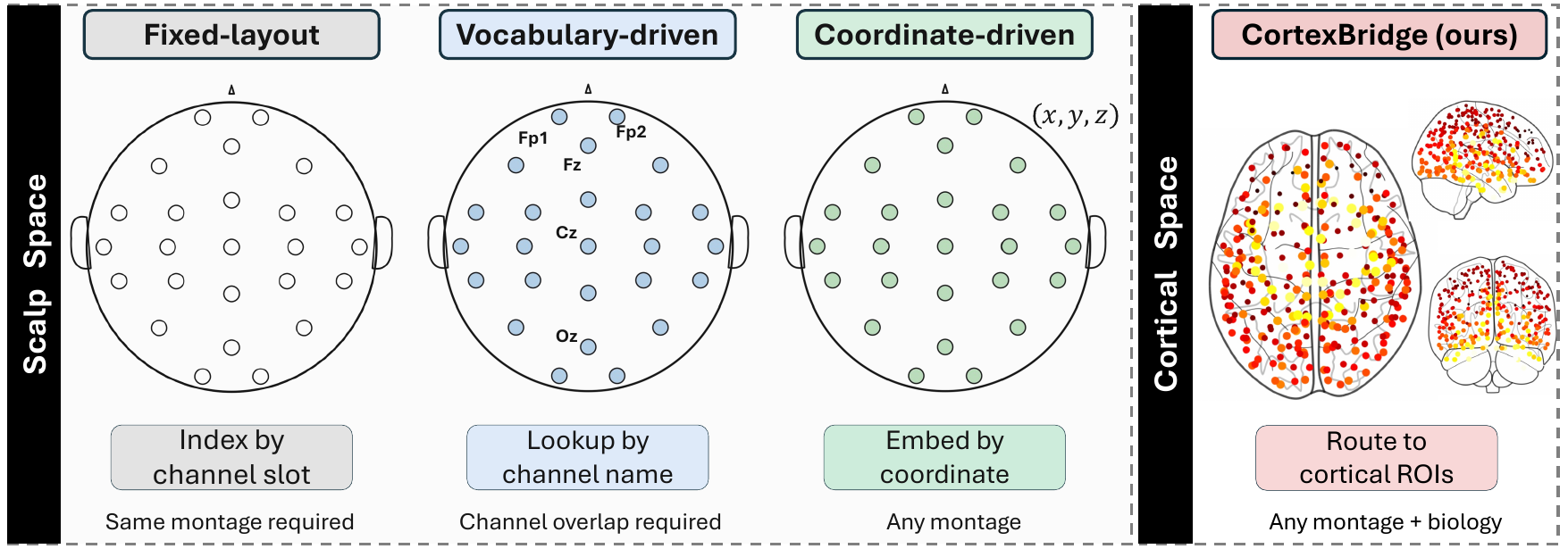}
\caption{Common interfaces for heterogeneous EEG montages. Fixed-layout models assume a consistent channel arrangement, vocabulary-driven models rely on predefined channel identities, and coordinate-driven models accommodate unseen electrode locations but remain in scalp space. \method{} instead maps arbitrary montages into a shared atlas-indexed cortical space.}
\label{fig:motivation}
\end{figure}

\paragraph{Montage interfaces.}
Existing models handle electrode layouts in several ways (Figure~\ref{fig:motivation}). Fixed-layout methods assume consistent channel slots across recordings, while vocabulary-driven methods identify electrodes from a predefined channel set. Coordinate-driven models such as NeurIPT and REVE instead encode electrode positions directly and can therefore accommodate previously unseen locations~\citep{fang2025neuript,ouahidi2025reve}. However, all three approaches ultimately retain a scalp-indexed representation. \method{} differs by mapping each montage to the same atlas-indexed cortical space before interfacing with the pretrained backbone.

\paragraph{Cortical representations.}
Channel selection and interpolation can reconcile input dimensions across montages but remain entirely in sensor space. EEG source imaging instead estimates cortical activity using a forward model and regularized inverse operator~\citep{michel2019eegsource,pascualmarqui2002sloreta}. \method{} serves a different purpose: it uses cortical anatomy as a representation prior and learns electrode-to-ROI routing directly for downstream adaptation. The Schaefer atlas and Yeo networks provide a common cortical organization~\citep{schaefer2018local,yeo2011organization}, while the resulting representation remains a learned latent space rather than a biophysical source estimate.

\begin{figure}
  \centering
  \includegraphics[width=\linewidth]{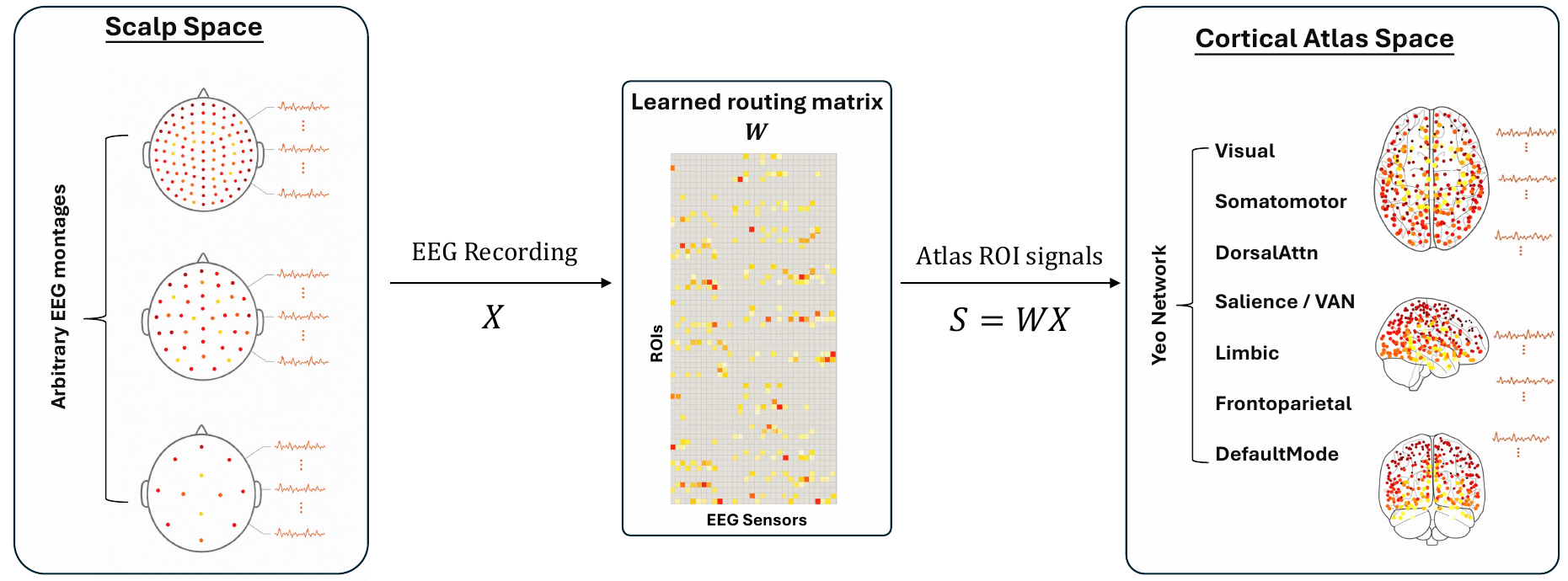}
  \caption{Overview of \method. The adapter maps montage-specific EEG to atlas ROIs, models cortical interactions in a latent space, and projects the result to the input of a frozen foundation model.}
  \label{fig:overview}
\end{figure}

\section{Method}

\subsection{Coordinate-aware cortical transport}

Let $\mathbf{X}\in\mathbb{R}^{B\times C\times T}$ contain $B$ EEG trials, $C$ electrodes, and $T$ samples. As illustrated in Figure~\ref{fig:overview}, \method{} maps the montage-specific channel dimension $C$ to a fixed set of $R$ atlas ROIs. Electrode coordinates are $\mathbf{P}\in\mathbb{R}^{C\times3}$, and atlas centroids are $\mathbf{Q}\in\mathbb{R}^{R\times3}$. Each electrode is summarized by its temporal mean, standard deviation, root mean square (RMS), line length, and four spectral power ratios. A small network combines these statistics with an embedding of $\mathbf P$, while the same position network embeds $\mathbf Q$. Multi-head cross-attention then produces
\begin{equation}
\ell_{b,h,r,c}=\frac{\langle \mathbf{q}_{h,r},\mathbf{k}_{b,h,c}\rangle}{\sqrt {d_h}}
-\operatorname{softplus}(\gamma_h)\|\widetilde{\mathbf Q}_r-\widetilde{\mathbf P}_c\|_2^2,
\end{equation}
where $h$ indexes attention heads and $d_h$ is the dimension of each head. Electrode and atlas coordinates are independently centered and divided by their mean pairwise distance. The second term provides a learned geometric bias rather than a fixed inverse solution.

For each head, we apply $K$ alternating row and column normalizations to the exponentiated logits, retain the largest $k$ entries for each ROI, and normalize each row. Averaging the $H$ head-specific matrices gives $\mathbf W=H^{-1}\sum_h\mathbf W^{(h)}\in\mathbb{R}^{B\times R\times C}$. Because every $\mathbf W^{(h)}$ is nonnegative and row stochastic, $\mathbf W$ is also nonnegative and row stochastic. The atlas signals are
\begin{equation}
\mathbf{S}_{b,r,t}=\sum_{c=1}^{C}W_{b,r,c}\mathbf{X}_{b,c,t}.
\end{equation}
Each ROI signal is therefore a convex mixture of electrode signals. A column-balance penalty discourages the routing from relying on only a small subset of electrodes.

\subsection{Deeper cortical latent space}

A single atlas projection does not model interactions among cortical regions. We construct row-normalized Gaussian kernels $\mathbf A^{(s)}\in\mathbb{R}^{R\times R}$ from distances between atlas centroids at three spatial scales. Learned weights combine the resulting short-, medium-, and long-range diffusion:
\begin{equation}
\overline{\mathbf S}=\sum_s \operatorname{softmax}(\boldsymbol\alpha)_s\mathbf A^{(s)}\mathbf S.
\end{equation}
Learned latent queries define a row-stochastic matrix $\mathbf U\in\mathbb{R}^{L\times R}$ that pools the $R$ ROIs into $L$ latent channels. A sigmoid gate $g_{\mathrm{lat}}$ combines the direct and diffused representations:
\begin{equation}
\mathbf Z=\operatorname{Norm}\left(\mathbf U\mathbf S+
g_{\mathrm{lat}}(\mathbf U\mathbf S)\odot \mathbf U\overline{\mathbf S}\right).
\end{equation}
This separates atlas alignment from the representation width required by a foundation model. We use $R=300$, $L=64$, four heads, and $k=8$ in all experiments.

\subsection{Foundation model adaptation}

The raw EEG and cortical representation are projected to the channel width required by each backbone. The resulting foundation-model input is
\begin{equation}
\mathbf X^{\mathrm{FM}}=P_{\mathrm{raw}}(\mathbf X)+
\tanh(\boldsymbol\eta)\odot P_{\mathrm{ctx}}(\mathbf Z),
\end{equation}
where $P_{\mathrm{raw}}$ and $P_{\mathrm{ctx}}$ are $1\times1$ projections and $\boldsymbol\eta$ is a channel-wise residual gate. We initialize $\boldsymbol\eta$ to zero, giving the raw and cortical models the same input when the cortical branch is introduced. The same interface is used for LaBraM, EEGPT, and CBraMod. We freeze each backbone and optimize the projections, \method, and the classification head.

Training combines the task loss with a column-balance term that discourages a small channel subset from receiving all routing mass. The balance weight is fixed across datasets. Top-$k$ selection already limits entropy, so no additional entropy penalty is used. Full objective and implementation details are provided in Appendix~\ref{app:repro}.

\section{Experiments}

\paragraph{Datasets.}
We evaluate on five public datasets from MOABB~\citep{jayaram2018moabb}, spanning motor imagery, P300, and SSVEP paradigms. Liu2024 contains left- and right-hand motor imagery from 50 acute stroke patients~\citep{liu2024stroke}. BNCI2014-008 contains visual P300 data from eight participants with amyotrophic lateral sclerosis~\citep{riccio2013attention}, while BNCI2015-003 uses an auditory P300 spelling task~\citep{schreuder2011listen}. Nakanishi2015 is a 12-class SSVEP dataset~\citep{nakanishi2015comparison}, and Huebner2017 contains visual P300 spelling data~\citep{huebner2017learning}. None of these datasets appears in the published pretraining sets of EEGPT, LaBraM, or CBraMod~\citep{jiang2024labram,wang2024eegpt,wang2024cbramod}.

\paragraph{Preprocessing and protocol.}
We use the event-specific windows defined by MOABB, apply a 0.5--45 Hz band-pass filter, and resample all datasets to 200 Hz. Additional model-specific resampling and padding are applied when required by each backbone's patch structure. Normalization statistics are computed from training subjects only. We use subject-grouped three-fold cross-validation, with test subjects held out from both training and validation. Balanced accuracy is reported for all experiments. Dataset details are provided in Appendix~\ref{app:data}.

\paragraph{Baselines and training.}
We evaluate EEGPT, CBraMod, and LaBraM on all five datasets. The raw baseline uses a learned channel projection, while \method{} augments this interface with the cortical path. Paired runs use identical subject folds, initialization, classifier, and training schedule. \method{} adds approximately 0.10M trainable parameters. Full training and implementation details are provided in Appendix~\ref{app:repro}.

\section{Results and Discussion}

\begin{table}[t]
\centering
\small
\setlength{\tabcolsep}{3.0pt}
\renewcommand{\arraystretch}{1.08}
\caption{Balanced accuracy (\%) under subject-grouped three-fold cross-validation. Blue cells use \method{}; arrows show the change from the corresponding raw interface. Bold marks the best result in each row.}
\label{tab:main}
\resizebox{\linewidth}{!}{%
\begin{tabular}{l cc cc cc}
\toprule
& \multicolumn{2}{c}{EEGPT} & \multicolumn{2}{c}{CBraMod} & \multicolumn{2}{c}{LaBraM} \\
\cmidrule(lr){2-3}\cmidrule(lr){4-5}\cmidrule(lr){6-7}
Dataset & Raw & \cellcolor{rowblue}+\method{} & Raw & \cellcolor{rowblue}+\method{} & Raw & \cellcolor{rowblue}+\method{} \\
\midrule
BNCI2014-008 & 55.49 & \cellcolor{rowblue}54.53\,\loss{0.96}
& 64.92 & \cellcolor{rowblue}\textbf{65.68}\,\gain{0.76}
& 50.43 & \cellcolor{rowblue}52.16\,\gain{1.72} \\
BNCI2015-003 & 55.84 & \cellcolor{rowblue}56.57\,\gain{0.72}
& 57.43 & \cellcolor{rowblue}\textbf{58.18}\,\gain{0.75}
& 50.62 & \cellcolor{rowblue}51.48\,\gain{0.86} \\
Nakanishi2015 & 61.67 & \cellcolor{rowblue}65.06\,\gain{3.40}
& 56.48 & \cellcolor{rowblue}\textbf{69.51}\,\gain{13.02}
& 15.56 & \cellcolor{rowblue}16.23\,\gain{0.68} \\
Liu2024 & 51.88 & \cellcolor{rowblue}52.50\,\gain{0.62}
& 54.35 & \cellcolor{rowblue}\textbf{55.20}\,\gain{0.85}
& 49.56 & \cellcolor{rowblue}49.89\,\gain{0.33} \\
Huebner2017 & 83.74 & \cellcolor{rowblue}\textbf{83.94}\,\gain{0.20}
& 78.11 & \cellcolor{rowblue}79.00\,\gain{0.89}
& 51.63 & \cellcolor{rowblue}51.52\,\loss{0.11} \\
\midrule
Average & 61.72 & \cellcolor{rowblue}62.52\,\gain{0.80}
& 62.26 & \cellcolor{rowblue}\textbf{65.51}\,\gain{3.26}
& 43.56 & \cellcolor{rowblue}44.26\,\gain{0.70} \\
\bottomrule
\end{tabular}}
\end{table}

\paragraph{Transfer across foundation models.}
\method{} improves 13 of the 15 dataset and model pairs in Table~\ref{tab:main}. Average balanced accuracy increases by 0.80\% for EEGPT, 3.26\% for CBraMod, and 0.70\% for LaBraM. CBraMod improves on all five datasets, while EEGPT and LaBraM improve on four. LaBraM gives the lowest absolute accuracy, which may reflect the difficulty of transferring its channel-wise tokenization and predefined channel interface to unseen montages~\citep{jiang2024labram}. Nevertheless, \method{} improves four of the five LaBraM results, showing that cortical alignment remains beneficial under this stronger interface constraint.

\paragraph{Transfer across EEG paradigms.}
The largest gain occurs on Nakanishi2015, where \method{} improves CBraMod by 13.02\% and EEGPT by 3.40\% on 12-class SSVEP. This result suggests that mapping sparse scalp measurements to a structured cortical representation is particularly useful for its eight-channel montage. All three models improve on BNCI2015-003 auditory P300, and smaller gains extend to Liu2024 motor imagery. The two decreases are limited to 0.96\% for EEGPT on BNCI2014-008 and 0.11\% for LaBraM on Huebner2017. The gains across motor imagery, P300, and SSVEP show that the learned routing is not limited to one EEG paradigm.

\begin{figure}[t]
  \centering
  \begin{subfigure}[t]{\linewidth}
    \centering
    \includegraphics[width=\linewidth]{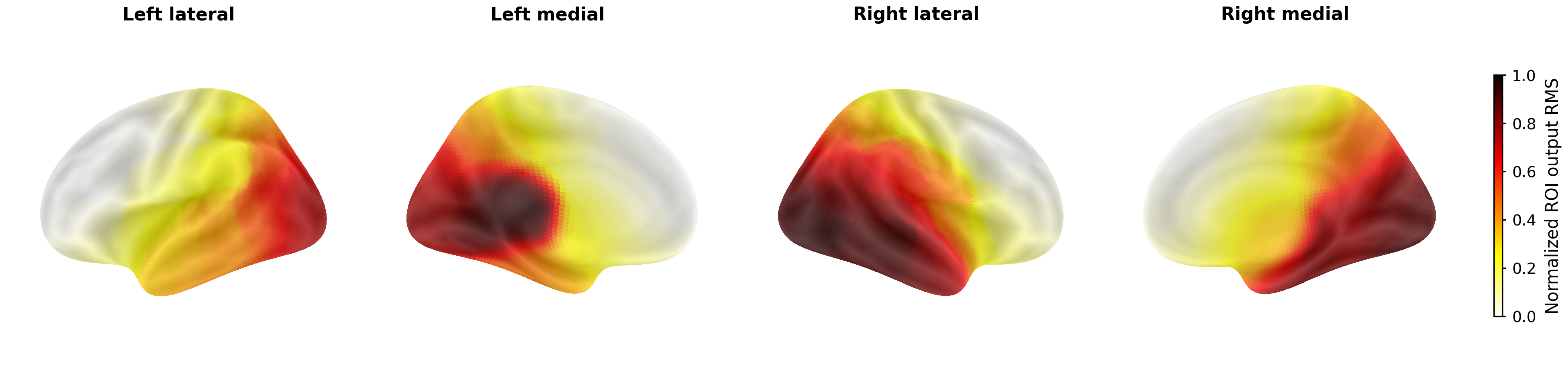}
    \caption{BNCI2015-003 auditory P300.}
  \end{subfigure}

  \vspace{2pt}
  \begin{subfigure}[t]{\linewidth}
    \centering
    \includegraphics[width=\linewidth]{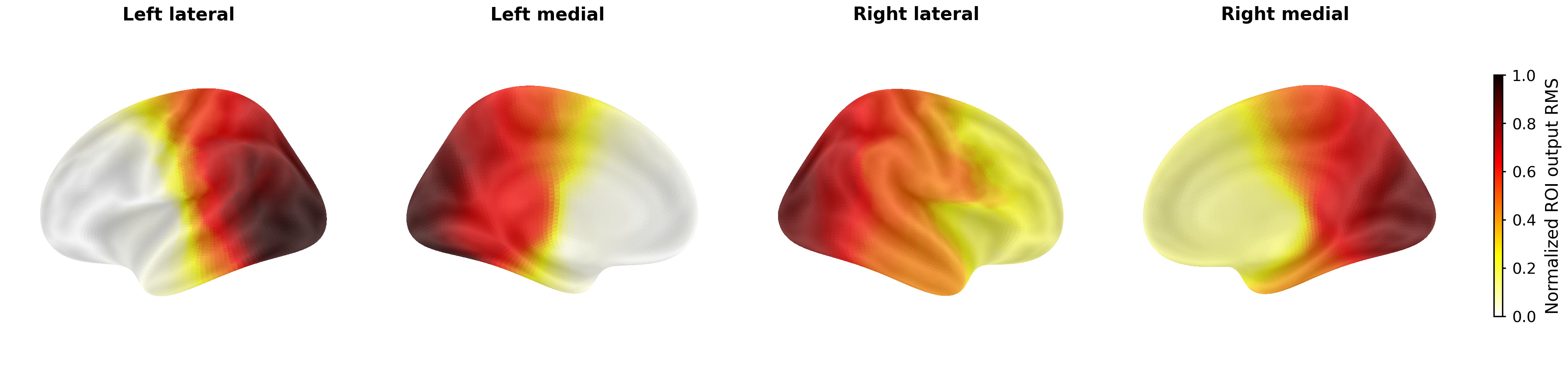}
    \caption{Nakanishi2015 SSVEP.}
  \end{subfigure}
  \caption{Atlas representations learned by CBraMod with \method. ROI-output RMS is normalized within each task using the 5th and 98th percentiles and clipped to $[0,1]$; colors indicate relative spatial distributions.}
  \label{fig:task_maps}
\end{figure}

\paragraph{Task-dependent cortical representations.}
Figure~\ref{fig:task_maps} compares atlas representations learned from two eight-channel datasets using the same backbone and adapter. SSVEP is elicited by periodic visual stimulation, and its representation is more concentrated in regions associated with the Yeo Visual network. This network captures the locally coupled organization of visual cortex in the seven-network parcellation~\citep{schaefer2018local,yeo2011organization}. This correspondence suggests that \method{} retains task-relevant cortical organization. The maps show learned ROI representations rather than biophysical source estimates.

\section{Conclusion}

We introduced \method, a lightweight adapter that maps heterogeneous scalp EEG montages to an atlas-indexed cortical representation through signal- and coordinate-aware electrode-to-ROI routing. Across EEGPT, LaBraM, and CBraMod, \method{} improves 13 of 15 evaluations on five unseen datasets, with a maximum gain of 13.02\% on 12-class SSVEP. The learned ROI representations retain task-dependent spatial organization, with SSVEP concentrated in the Yeo Visual network. These results show that \method{} improves adaptation to unseen montages and provides a common, anatomically indexed interface across EEG foundation models.

\clearpage
\bibliographystyle{plainnat}
\bibliography{reference}

@inproceedings{jiang2024labram,
  title={Large Brain Model for Learning Generic Representations with Tremendous {EEG} Data in {BCI}},
  author={Jiang, Wei-Bang and Zhao, Li-Ming and Lu, Bao-Liang},
  booktitle={International Conference on Learning Representations},
  year={2024}
}

@inproceedings{wang2024eegpt,
  title={{EEGPT}: Pretrained Transformer for Universal and Reliable Representation of {EEG} Signals},
  author={Wang, Guangyu and Liu, Wenchao and He, Yuhong and Xu, Cong and Ma, Lin and Li, Haifeng},
  booktitle={Advances in Neural Information Processing Systems},
  volume={37},
  year={2024},
  doi={10.52202/079017-1239}
}

@inproceedings{wang2024cbramod,
  title={{CBraMod}: A Criss-Cross Brain Foundation Model for {EEG} Decoding},
  author={Wang, Jiquan and Zhao, Sha and Luo, Zhiling and Zhou, Yangxuan and Jiang, Haiteng and Li, Shijian and Li, Tao and Pan, Gang},
  booktitle={International Conference on Learning Representations},
  year={2025}
}

@article{michel2019eegsource,
  title={{EEG} Source Imaging: A Practical Review of the Analysis Steps},
  author={Michel, Christoph M. and Brunet, Denis},
  journal={Frontiers in Neurology},
  volume={10},
  pages={325},
  year={2019},
  doi={10.3389/fneur.2019.00325}
}

@article{pascualmarqui2002sloreta,
  title={Standardized Low-Resolution Brain Electromagnetic Tomography ({sLORETA}): Technical Details},
  author={Pascual-Marqui, Roberto D.},
  journal={Methods and Findings in Experimental and Clinical Pharmacology},
  volume={24},
  number={Suppl D},
  pages={5--12},
  year={2002}
}

@article{schaefer2018local,
  author={Schaefer, Alexander and Kong, Ru and Gordon, Evan M. and Laumann, Timothy O. and Zuo, Xi-Nian and Holmes, Avram J. and Eickhoff, Simon B. and Yeo, B. T. Thomas},
  title={Local-Global Parcellation of the Human Cerebral Cortex from Intrinsic Functional Connectivity {MRI}},
  journal={Cerebral Cortex},
  volume={28},
  number={9},
  pages={3095--3114},
  year={2018},
  doi={10.1093/cercor/bhx179}
}

@article{yeo2011organization,
  author={Yeo, B. T. Thomas and Krienen, Fenna M. and Sepulcre, Jorge and Sabuncu, Mert R. and Lashkari, Danial and Hollinshead, Marisa and Roffman, Joshua L. and Smoller, Jordan W. and Z{\"o}llei, Lilla and Polimeni, Jonathan R. and Fischl, Bruce and Liu, Hesheng and Buckner, Randy L.},
  title={The Organization of the Human Cerebral Cortex Estimated by Intrinsic Functional Connectivity},
  journal={Journal of Neurophysiology},
  volume={106},
  number={3},
  pages={1125--1165},
  year={2011},
  doi={10.1152/jn.00338.2011}
}

@article{jayaram2018moabb,
  title={{MOABB}: Trustworthy Algorithm Benchmarking for {BCI}s},
  author={Jayaram, Vinay and Barachant, Alexandre},
  journal={Journal of Neural Engineering},
  volume={15},
  number={6},
  pages={066011},
  year={2018},
  doi={10.1088/1741-2552/aadea0}
}

@article{liu2024stroke,
  title={An {EEG} Motor Imagery Dataset for Brain Computer Interface in Acute Stroke Patients},
  author={Liu, Haijie and Wei, Penghu and Wang, Haochong and Lv, Xiaodong and Duan, Wei and Li, Meijie and Zhao, Yan and Wang, Qingmei and Chen, Xinyuan and Shi, Gaige and Han, Bo and Hao, Junwei},
  journal={Scientific Data},
  volume={11},
  pages={131},
  year={2024},
  doi={10.1038/s41597-023-02787-8}
}

@article{riccio2013attention,
  title={Attention and {P300}-Based {BCI} Performance in People with Amyotrophic Lateral Sclerosis},
  author={Riccio, Angela and Simione, Luca and Schettini, Francesca and Pizzimenti, Alessia and Inghilleri, Maurizio and Olivetti Belardinelli, Marta and Mattia, Donatella and Cincotti, Febo},
  journal={Frontiers in Human Neuroscience},
  volume={7},
  pages={732},
  year={2013},
  doi={10.3389/fnhum.2013.00732}
}

@article{schreuder2011listen,
  title={Listen, You Are Writing! Speeding Up Online Spelling with a Dynamic Auditory {BCI}},
  author={Schreuder, Martijn and Rost, Thomas and Tangermann, Michael},
  journal={Frontiers in Neuroscience},
  volume={5},
  pages={112},
  year={2011},
  doi={10.3389/fnins.2011.00112}
}

@article{nakanishi2015comparison,
  title={A Comparison Study of Canonical Correlation Analysis Based Methods for Detecting Steady-State Visual Evoked Potentials},
  author={Nakanishi, Masaki and Wang, Yijun and Wang, Yu-Te and Jung, Tzyy-Ping},
  journal={PLOS ONE},
  volume={10},
  number={10},
  pages={e0140703},
  year={2015},
  doi={10.1371/journal.pone.0140703}
}

@article{huebner2017learning,
  title={Learning from Label Proportions in Brain-Computer Interfaces: Online Unsupervised Learning with Guarantees},
  author={H{\"u}bner, David and Verhoeven, Thibault and Schmid, Konstantin and M{\"u}ller, Klaus-Robert and Tangermann, Michael and Kindermans, Pieter-Jan},
  journal={PLOS ONE},
  volume={12},
  number={4},
  pages={e0175856},
  year={2017},
  doi={10.1371/journal.pone.0175856}
}

@article{lawhern2018eegnet,
  title={{EEGNet}: A Compact Convolutional Neural Network for {EEG}-Based Brain--Computer Interfaces},
  author={Lawhern, Vernon J. and Solon, Amelia J. and Waytowich, Nicholas R. and Gordon, Stephen M. and Hung, Chou P. and Lance, Brent J.},
  journal={Journal of Neural Engineering},
  volume={15},
  number={5},
  pages={056013},
  year={2018},
  doi={10.1088/1741-2552/aace8c}
}

@article{song2022eeg,
  title={{EEG} Conformer: Convolutional Transformer for {EEG} Decoding and Visualization},
  author={Song, Yonghao and Zheng, Qingqing and Liu, Bingchuan and Gao, Xiaorong},
  journal={IEEE Transactions on Neural Systems and Rehabilitation Engineering},
  volume={31},
  pages={710--719},
  year={2022},
  doi={10.1109/TNSRE.2022.3230250}
}

@article{kostas2021bendr,
  title={{BENDR}: Using Transformers and a Contrastive Self-Supervised Learning Task to Learn from Massive Amounts of {EEG} Data},
  author={Kostas, Demetres and Aroca-Ouellette, Stephane and Rudzicz, Frank},
  journal={Frontiers in Human Neuroscience},
  volume={15},
  pages={653659},
  year={2021},
  doi={10.3389/fnhum.2021.653659}
}

@article{chien2022maeeg,
  title={{MAEEG}: Masked Auto-Encoder for {EEG} Representation Learning},
  author={Chien, Hsiang-Yun Sherry and Goh, Hanlin and Sandino, Christopher M. and Cheng, Joseph Y.},
  journal={arXiv preprint arXiv:2211.02625},
  year={2022},
  note={NeurIPS 2022 Workshop on Learning from Time Series for Health}
}

@inproceedings{mohammadi2024eeg2rep,
  title={{EEG2Rep}: Enhancing Self-Supervised {EEG} Representation through Informative Masked Inputs},
  author={Mohammadi Foumani, Navid and Mackellar, Geoffrey and Ghane, Soheila and Irtza, Saad and Nguyen, Nam and Salehi, Mahsa},
  booktitle={Proceedings of the 30th ACM SIGKDD Conference on Knowledge Discovery and Data Mining},
  pages={5544--5555},
  year={2024},
  doi={10.1145/3637528.3671600}
}

@inproceedings{yi2023learning,
  title={Learning Topology-Agnostic {EEG} Representations with Geometry-Aware Modeling},
  author={Yi, Ke and Wang, Yansen and Ren, Kan and Li, Dongsheng},
  booktitle={Advances in Neural Information Processing Systems},
  volume={36},
  pages={53875--53891},
  year={2023}
}

@inproceedings{yang2023biot,
  title={{BIOT}: Biosignal Transformer for Cross-Data Learning in the Wild},
  author={Yang, Chaoqi and Westover, M. Brandon and Sun, Jimeng},
  booktitle={Advances in Neural Information Processing Systems},
  volume={36},
  year={2023}
}

@inproceedings{fang2025neuript,
  title={{NeurIPT}: Foundation Model for Neural Interfaces},
  author={Fang, Zitao and Li, Chenxuan and Zhou, Hongting and Yu, Shuyang and Du, Guodong and Qasem, Ashwaq and Lu, Yang and Li, Jing and Zhang, Junsong and Goh, Sim Kuan},
  booktitle={Advances in Neural Information Processing Systems},
  year={2025},
  note={arXiv:2510.16548}
}

@inproceedings{ouahidi2025reve,
  title={{REVE}: A Foundation Model for {EEG}---Adapting to Any Setup with Large-Scale Pretraining on 25,000 Subjects},
  author={El Ouahidi, Yassine and Lys, Jonathan and Th{\"o}lke, Philipp and Farrugia, Nicolas and Pasdeloup, Bastien and Gripon, Vincent and Jerbi, Karim and Lioi, Giulia},
  booktitle={Advances in Neural Information Processing Systems},
  year={2025}
}

@article{hong2025samba,
  title={SAMBA: Toward a Long-Context {EEG} Foundation Model via Spatial Embedding and Differential Mamba},
  author={Hong, Jiazhen and Mackellar, Geoffrey and Ghane, Soheila},
  journal={arXiv preprint arXiv:2511.18571},
  year={2025}
}

\appendix
\section{Dataset details}\label{app:data}
\begin{center}
\small
\begin{tabular}{lrrrrl}
\toprule
Dataset & Subjects & Ch. & Classes & Native Hz & Task \\
\midrule
Liu2024 & 50 & 29 & 2 & 500 & Motor imagery \\
BNCI2014-008 & 8 & 8 & 2 & 256 & P300, amyotrophic lateral sclerosis \\
BNCI2015-003 & 10 & 8 & 2 & 256 & Auditory P300 \\
Nakanishi2015 & 9 & 8 & 12 & 256 & SSVEP \\
Huebner2017 & 13 & 31 & 2 & 1000 & Visual P300 \\
\bottomrule
\end{tabular}
\end{center}

\paragraph{Liu2024.}
This dataset contains left- and right-hand grip motor imagery from 50 acute stroke patients. MOABB provides 29 EEG channels recorded at 500~Hz, with 20 trials per class and four-second task epochs~\citep{liu2024stroke}.

\paragraph{BNCI2014-008.}
Eight participants with amyotrophic lateral sclerosis performed a visual row--column P300 spelling task. The binary target and non-target responses were recorded from eight channels at 256~Hz~\citep{riccio2013attention}.

\paragraph{BNCI2015-003.}
Ten participants used an auditory P300 speller based on spatial auditory cues. The dataset contains binary target and non-target responses from eight channels sampled at 256~Hz~\citep{schreuder2011listen}.

\paragraph{Nakanishi2015.}
The MOABB release contains nine participants viewing 12 joint frequency--phase modulated SSVEP targets. Each class has 15 trials recorded from eight channels at 256~Hz~\citep{nakanishi2015comparison}.

\paragraph{Huebner2017.}
Thirteen participants performed a visual matrix-speller task. MOABB provides binary target and non-target responses from 31 channels recorded at 1000~Hz~\citep{huebner2017learning}.

\section{Implementation details}\label{app:repro}
We use 300 Schaefer atlas regions, 64 latent channels, four attention heads, four alternating normalization steps, and top-$k$ routing with $k=8$. The objective is
\begin{equation}
\mathcal L=\mathcal L_{\mathrm{task}}+10^{-2}\frac{1}{B}\sum_b
\operatorname{Var}_{c}\left(\sum_r W_{b,r,c}\right).
\end{equation}
For each fold, we first train the raw projection and classification head for two epochs. The resulting state initializes both the raw and \method{} conditions, which are then trained for up to five epochs. We use AdamW with weight decay $10^{-2}$, a learning rate of $10^{-4}$ for shared parameters, and $5\times10^{-4}$ for the cortical path. The pretrained encoder remains frozen. Validation subjects are selected only from the training fold, and model selection uses validation balanced accuracy. All experiments use seed 2026.

\section{Limitations}\label{app:limits}
Our evaluation covers five public datasets, three EEG paradigms, and three foundation models. Future work will extend \method{} to more recording systems, participant populations, and model architectures. Evaluation on larger mixed-montage collections and different atlas resolutions will further establish its generality. Theoretical analysis of the learned cortical mapping is another direction for future work.

\end{document}